\documentclass[10pt,twocolumn,letterpaper]{article}

\usepackage{wacv}              

\usepackage{graphicx}
\usepackage{amsmath}
\usepackage{amssymb}
\usepackage{booktabs}
\usepackage{algorithm}
\usepackage{algpseudocode}
\usepackage{adjustbox}

\usepackage[pagebackref,breaklinks,colorlinks]{hyperref}

\usepackage[capitalize]{cleveref}
\crefname{section}{Sec.}{Secs.}
\Crefname{section}{Section}{Sections}
\Crefname{table}{Table}{Tables}
\crefname{table}{Tab.}{Tabs.}

\def\wacvPaperID{2005} 
\def\confName{WACV}
\def\confYear{2024}

\usepackage{mathtools}
\usepackage{amsthm}
\usepackage{subdef}
\usepackage{subcaption}
\usepackage{multirow}
\newcommand{\ours}{\textsc{DeDiER}}

\usepackage[accsupp]{axessibility} 
\usepackage{xcolor}

\usepackage{colortbl}
\newcommand*\ouralgo{\ours} 

\begin{document}

\title{Using Early Readouts to Mediate Featural Bias in Distillation}

\author{
Rishabh Tiwari\\
{ Google Research, India}\\
{\tt\small rishabhtiwari@google.com}
\and
Durga Sivasubramanian\\
{IIT Bombay}\\
{\tt\small durgas@cse.iitb.ac.in}
\and
Anmol Mekala\\
{ University of Massachusetts Amherst}\\
{\tt\small amekala@umass.edu}
\and
Ganesh Ramakrishnan\\
{IIT Bombay}\\
{\tt\small ganesh@cse.iitb.ac.in}
\and
Pradeep Shenoy\\
{Google Research, India}\\
{\tt\small shenoypradeep@google.com}
}
\maketitle

\begin{abstract}
   Deep networks tend to learn spurious feature-label correlations in real-world supervised learning tasks. This vulnerability is aggravated in distillation, where a student model may have lesser representational capacity than the corresponding teacher model. Often, knowledge of specific spurious correlations is used to reweight instances \& rebalance the learning process. We propose a novel early readout mechanism whereby we attempt to predict the label using representations from earlier network layers. We show that these early readouts automatically identify problem instances or groups in the form of confident, incorrect predictions. Leveraging these signals to modulate the distillation loss on an instance level allows us to substantially improve not only group fairness measures across benchmark datasets, but also overall accuracy of the student model.
  We also provide secondary analyses that bring insight into the role of feature learning in supervision and distillation.
\end{abstract}

\section{Introduction}

Deep networks trained via supervision tend to preferentially learn simple features with weak label correlations~\cite{hermann2020shapes, geirhos2020shortcut, shah2020pitfalls}. This weakness is significantly magnified in real-world applications where limited data or sampling biases may introduce spurious correlations between unrelated features and desired labels. As an example, the Waterbirds benchmark~\cite{sagawa2020group} challenges DNNs' dependence on habitat or background cues for classifying bird species.  These challenges are significantly worsened in distillation, where both teacher model biases and student model capacity limitations may worsen fairness measures in the student~\cite{lukasik2021teacher,ahn2022knowledge, du2021robustness}.

\begin{figure*}[!ht]
\begin{subfigure}[h]{0.53\linewidth}
\includegraphics[width=\linewidth]{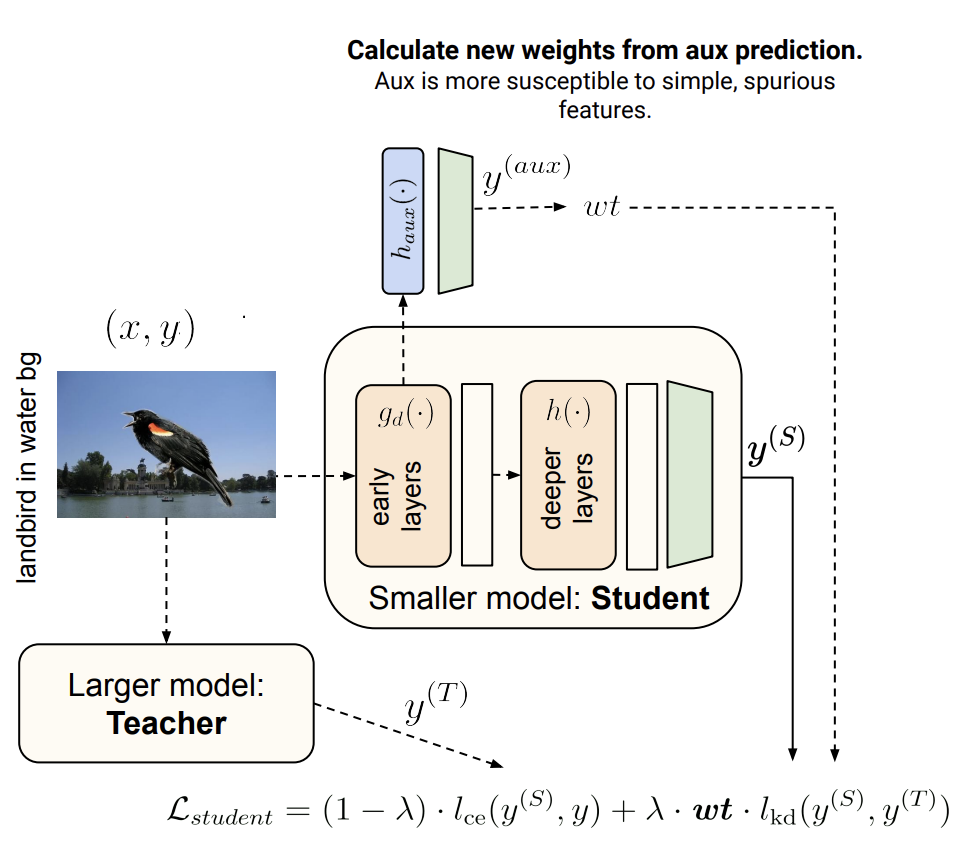}
\caption{A visual illustration of \ours{}}
\label{fig:method-diagram}
\end{subfigure}
\hfill
\begin{subfigure}[h]{0.45\linewidth}
\includegraphics[width=\linewidth]{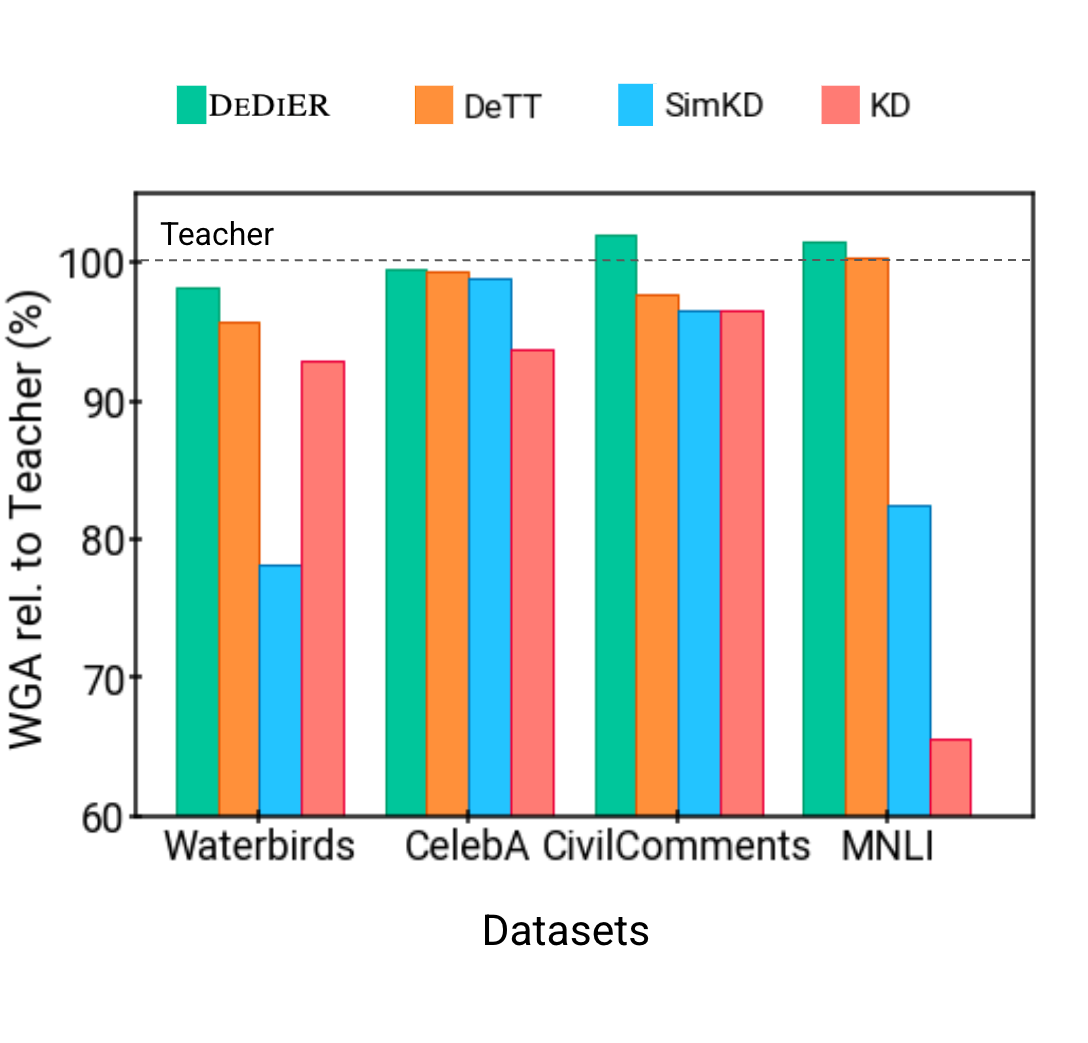}
\caption{Worst-group performances of the different KD methods.}
\label{fig:performance-comparison}
\end{subfigure}%
\caption{(a) We use predictions from an auxiliary layer applied on top of early features to determine the weights for the distillation loss. Errors from the readouts are disproportionately from learned spurious features. (b) Comparison of Worst Group Accuracies (WGA) relative to that of the Teacher's. \ours{} is best in being able to match the Teacher's WGA.}
\end{figure*}

We focus on the problem of learning debiased student models in a distillation setting. Bias is typically studied by tracking model performance by relevant subgroups of data, \eg, groups where a spurious attribute is in agreement with, or in conflict with the true label. Directly optimizing such measures of bias (\eg, Group DRO~\cite{sagawa2020group}) is limited by the need for annotations of group knowledge or feature dimensions known to be spuriously correlated. Recent work instead uses errors of a pre-trained model as a proxy for identifying problem instances, and a subsequent round of training with upweighted losses for those instances~\cite{liu2021just,lee2023debiased}. These methods need to carefully titrate early stopping criteria for identifying erroneous instances, since DNNs typically overfit on training data and rapidly drive training error rates to zero.

We  propose \ours\ (\textbf{De}biased \textbf{Di}stillation using \textbf{E}arly \textbf{R}eadouts) wherein we introduce two key innovations for debiasing distilled student models:  \textbf{a) Early readouts} -- a new source of differential information about spurious feature correlations at an instance level, and \textbf{b) Confidence weighting} -- which allows for fine-grained adjustments to distillation on a per-instance level, graded by prediction confidence. Our primary insight is that spurious features are often learned at earlier levels of a neural network~\cite{hermann2020shapes,hu2018gather,pmlr-v202-tiwari23a}; thus, prediction errors at early layers can provide significant information about feature bias in the network. Indeed, we show that ``reading out'' (\ie, predicting with a linear decoder) instance labels from earlier representations disproportionately errs on instances that defy known spurious correlations (\cref{fig:early_vs_late} (a, c)), indicating overdependence on those features. Further, the readouts are often \textit{confidently} wrong on those conflicting instances as compared to other errors (\cref{fig:early_vs_late} (b, d)). This leads naturally into our proposal for bias-mitigated distillation: modulate the teacher signal by a function of early readout confidence margin (\cref{fig:method-diagram}). A significant strength of our proposal is that it can be easily attached to any standard model training procedure. This means that the early readouts provide information about the specific model being trained, and also evolve through the training procedure, unlike previous approaches which depend on static identification of instances by a separate pre-trained model. 



Summing up, we make the following contributions:
\begin{itemize}
    \item We propose \textit{early readouts} as a novel source of information for identifying the risk of bias at the instance level. Our approach does not rely on any specialized knowledge of the dataset, such as group membership or spurious attribute values in instances. 
    \item We propose a flexible, margin-based method for reweighing the teacher-matching loss in distillation, using these early readouts, to significantly mitigate bias in student models. Our approach outperforms SOTA on not just fairness measures but also overall accuracy on well-studied debiasing benchmarks (\cref{fig:performance-comparison}).
    \item We demonstrate the generality of our instance-weighting scheme by evaluating on 4 different datasets having different group biases and compositions.
\end{itemize}

\section{Related Work}

\subsection{Bias \& Group-Fairness}
Machine learning models are typically trained with a goal of maximizing average performance. However, there can be hidden biases present in the data which can inadvertently be perpetuated or even amplified by ML models~\cite{Buolamwini2018GenderSI, Jurgens2017IncorporatingDV}. In particular, recent literature highlights the tendency of DNNs to focus on easy-to-learn features (``simplicity bias'', see \eg,~\cite{geirhos2020shortcut, nam2020learning, teney2022evading}). This inherent weakness of DNNs can magnify existing spurious correlations in the data by preferentially learning those features. Various algorithms for distributionally robust optimization (DRO) have been proposed to address this issue~\cite{Hovy2015TaggingPC, hashimoto2018fairness, duchi2018learning}, but their formulation is too conservative. More recent work~\cite{sagawa2020group, soma2022optimal} proposed a general formulation which explicitly optimizes over group labels. Setlur {\em et. al.}~\cite{setlur2022bitrateconstrained} build on these ideas without using group labels and without sacrificing overall accuracy. Other work proposes simple repeated training recipes~\cite{liu2021just} that aim to reduce errors from previous trained models as a proxy for debiasing models. 




\subsection{Knowledge Distillation}
Knowledge distillation (KD) uses a larger \textit{teacher} model to help small \textit{student} models learn more than they would while using cross-entropy loss over hard labels. This network compression is most commonly achieved by making the student mimic the teacher's final output layer logits~\cite{hinton2015distilling}. Other variants~\cite{heo2019comprehensive} also distil the teacher's intermediate feature maps for a more thorough transfer of the teacher's knowledge. SimKD~\cite{chen2022knowledge} transplants the teacher's discriminative head to the student in addition to performing feature distillation, thus aiming for a closer matching of the teacher performance.

\subsection{Bias and Distillation}
The robustness and bias of compressed models has been investigated in several works. Pruning-based compression has been observed to worsen model performance on underrepresented groups~\cite{hooker2020characterising} and worsen OOD performance \cite{du2021robustness}. KD-trained small models have been observed to have amplified bias~\cite{ahn2022knowledge} and worsened OOD performance~\cite{du2021robustness}. 
The picture that emerges from these works is that smaller models learnt via different compression methods such as pruning, KD, \etc., tend to rely on spurious correlations while aiming to match the average accuracy of larger models. Although the larger models are less dependent on spurious correlations, this knowledge is not transferred completely during KD; this is the problem that we aim to solve.

Mitigating of bias in distillation has been previously attempted by the softening of teacher labels based on sample hardness~\cite{du2021robustness}, data augmentation via mixup on protected characteristics~\cite{ahn2022knowledge}, and transplanting (robust) teacher layers~\cite{lee2023debiased}. Du {\em et. al.}~\cite{du2021robustness} employ an ensemble of models of various sizes to decide sample hardness, whereas \cite{ahn2022knowledge} depends on the availability of specific group annotations;  thus these approaches are inadequate for our setting. Closest to our setting is \cite{lee2023debiased}; we compare against and significantly outperform this approach.

\section{Preliminaries}



\begin{figure}[t]
\centering
\hspace{-0.6cm}
\includegraphics[width=\linewidth]{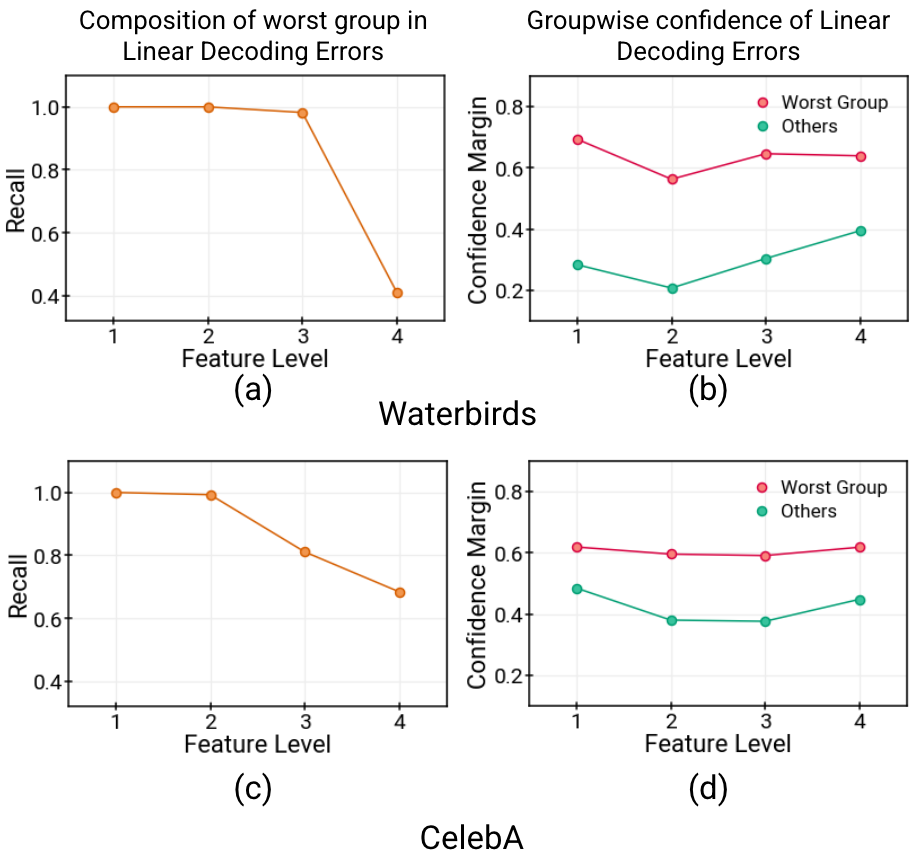}
\caption{Early readout errors recall worst group instances (left) and worst group readouts are more confident, across layers (right). We measure linear decoding error and confidence margins at each layer, after 1 epoch. (a, c): We observe that nearly all worst group instances are misclassified ($\sim 100$\% recall) with more recall in earlier layers. (b, d): show that error instances from minority groups have significantly higher confidence margin compared to other groups. See text for more details.}
\label{fig:early_vs_late}
\end{figure}

We aim to build a classifier model $f(\cdot)$ using training data $D = {(x_i, y_i) \mid i \in (1,\cdots,n) }$. Here, $x_i \in \Xcal$ denotes the input of the $i^{th}$ instance, and $y_i \in \Ycal$ the corresponding label. The model $f$ is parameterized by $\theta \in \Theta$, and our goal is to learn a mapping $f_{\theta}: \Xcal \rightarrow \Ycal$. We define the 0-1 loss function as 
$l_{0-1}(x, y ; \theta) = \bone[f_\theta(x) \neq y]$ representing the error incurred. In a standard classification setting, we seek to minimize $\EE \left[\ell_{0-1}(x, y; \theta)\right]$, \ie, the 0-1 loss over the entire dataset.

\noindent \textbf{Knowledge Distillation:} \label{sec:kd} Suppose we have access to a pre-trained model (teacher) that generates logits denoted as $y^{(T)} = \Tcal(\xb)$. Then a new model (student) could be trained using a ``teacher matching'' objective that involves minimizing the KL-divergence between the student's logits $y^{(S)}$ and those of the teacher $y^{(T)}$ \cite{hinton2015distilling}. The training objective typically used to train a student model is:
\begin{align}
      \mathcal{L}_{student} = \sum_D \left((1-\lambda)l_{ce} + \lambda l_{kd}\right)
    \label{eqdis1}
\end{align}
where $l_{kd} = \tau^2 KL\big(y^{(S)},y^{(T)}\big)$ is the teacher-matching loss, $\tau$ is a temperature parameter that controls the softening of the KL-divergence term, and $l_{ce} = H\big(y^{(S)},y\big)$ is a supervised learning loss on the model's predictions $y^{(S)}$ compared to the true labels $y$. $H$ is typically cross-entropy loss, a tractable relaxation of the 0-1 loss mentioned above. Finally, $\lambda$ is a hyperparameter controlling the contribution of the above two loss components.

\noindent \textbf{Groups and spurious correlations:} We assume that the dataset $D$ consists of groups $g \in G$, and that each data point in $D$ belongs to one of these groups in $G$. The groups are defined in terms of attribute value and label pairs where the attribute value may be spuriously correlated with the label in the supplied dataset $D$. As an example, in the Waterbirds dataset, the background of a given image (\eg, water) may be heavily associated with a specific category of foreground objects (waterbirds); however, this correlation is clearly incidental, not causal. Thus, the relevant groups for this dataset consist of different pairings of background and foreground objects. Continuing our Waterbirds example, landbirds on water are heavily misclassified, as the spurious feature (water) is in conflict with the true label. We refer to such groups interchangeably in the text as ``worst groups'' (groups with worst accuracy, typically) or ``minority groups'' (since such conflicting examples are in the minority in typical datasets).

\noindent \textbf{Group fairness:} Minimizing the overall 0-1 loss on a dataset does not necessarily guarantee group-wise fairness (\ie, comparable performance on all groups in the data); indeed, in practice, DNNs are heavily swayed by spurious correlations, leading to errors when the spurious feature's prediction is in conflict with the label. One way of addressing this is by defining a group fairness loss such as Group DRO~\cite{sagawa2020group}:  $\max_{g \in G} \EE \left[\ell_{0-1}(x, y; \theta) \mid g\right]$. By minimizing this \textit{worst group} accuracy, one can achieve equitable performance across groups. Alternatively, one could use the above loss as a measure of performance for comparing methods on their group fairness at test-time. Similar objectives can be framed for distillation as well.

\noindent \textbf{Fairness without group labels:} Directly optimizing a groupwise loss requires group information for each training instance, which may be limiting or even infeasible in practice. In addition to substantial labeling costs, a concern is that only some spurious correlations may be captured via explicit labeling. Instead, recent work uses model prediction errors to identify problematic instances, as a proxy for more generally identifying spurious correlations. JTT~\cite{liu2021just} trains a standard supervised classifier, collects instance errors from that classifier, and then trains a fresh classifier with a large fixed multiplier on the loss from those instances. The hope is that errors from the first model signal instances where spurious features contradict the label; by upweighting those instances, JTT shows improvements in group fairness measures even though no group information was used in training. DeTT \cite{lee2023debiased} uses a similar strategy for upweighting teacher loss for error instances in distillation, by using an early-stopped ERM model for identifying error instances. These approaches depend on proxy models trained via careful early stopping criteria, and only capture a fraction of the problematic instances in the dataset.

\section{Debiased Distillation with \ours}



We now describe \ours{}, a novel method for distilling knowledge from a debiased teacher without using group information in the training data. We identify \textit{early readouts}--label predictions from early network representations--as a novel signal for overdependence on spurious features, and leverage this signal into a graded adjustment of distillation loss on a per-instance level (\cref{fig:method-diagram}). Since these signals are directly accessible from the model being trained, rather than proxy models, we can develop a dynamic weighting scheme for distillation wherein our adjustments adapt to the model as it is being learned.

We first motivate the use of early readouts for automatically identifying problem instances, and more broadly, groups of instances that may suffer from spurious feature correlations (\cref{fig:early_vs_late}). We then show how \textit{confidence margins} from early readouts can be flexibly transformed into a graded weighting scheme for distillation (\cref{fig:upweighting}).

\subsection{Early Readouts for Precise Identification}

Our key insight is that early representations in the network provide significantly richer information about errors and spurious correlations, compared to the final predictions made by a classifier. This insight is based on previous work that suggests that DNNs are led astray by ``simple'' features (the so-called simplicity bias~\cite{shah2020pitfalls}); further, that simple features are typically learned early in the network stack, and spread throughout later layers~\cite{hermann2020shapes}. 

To illustrate this idea clearly, we trained a ResNet-18 ~\cite{he2016deep} model on the Waterbirds and CelebA datasets~\cite{wah2011cub,sagawa2020group,liu2015deep} for one epoch, and trained linear decoders that operate on the representation from each layer of the network. We examined how error, and error + confidence, can be used to differentiate minority group instances from other groups. We summarize our observations next: 

\noindent \textbf{Early errors signal worst groups:} In \cref{fig:early_vs_late} (a) we observe, how on the Waterbirds dataset, nearly all minority instances are incorrectly classified at earlier network layers ($\sim100$\% recall), although this error drops rapidly at the final layer even after a single epoch of training. This is already a substantial advantage over previous approaches such as JTT \& DeTT, that use error instances at the final layer of an ERM classifier to identify minority instances. The finding is replicated on the CelebA dataset (\cref{fig:early_vs_late} (c)). This also confirms the hypothesis that early readouts give a strong indication of the network's dependence on spuriously correlated features.

\noindent \textbf{Erroneous but confident:} Next we examine whether the \textit{confidence margin} associated with incorrectly labeled instances provide further discriminative information. For a probabilistic classifier output $\mathbf{p}=\{p_1,\cdots,p_K\}$, where $p_k$ denotes the probability of label $k$, we define the confidence margin as the difference between the top two values of $p_k$:
\begin{align}
p_{max} &= \max_{p_k\in \mathbf{p}} p_k \\
\mathtt{cm}(\mathbf{p}) &= p_{max} - \max_{p_k\in \mathbf{p}\backslash p_{max}} p_k
\label{eq:confmargin}
\end{align}
In \cref{fig:early_vs_late} (b, d) we present the average confidence margin on incorrectly classified instances, broken down into minority group and other groups (Waterbirds). We see that confidence margin further helps us discriminate between the two sets of erroneous instances, with minority group errors associated with significantly higher margin. Again, this differential is larger in earlier rather than later layers.

\noindent \textbf{Advantages of early readouts:} These results clearly bring out the following points: (a) Errors from early readouts, and in particular, the associated confidence margins, are significantly more informative than errors at final layers of the model (\eg, \cite{liu2021just,lee2023debiased}). (b) Since the readouts are a cheap way of monitoring network performance at an instance level, we can design \textit{dynamic} adjustment techniques that use readouts from the classifier being learned, throughout the course of its training, instead of a fixed upweighting of a pre-selected set of instances from a proxy model's output~\cite{liu2021just,lee2023debiased}. 
As we show in our results, this weighing scheme is more effective with earlier layers.

\subsection{Confidence-Based Weighting for KD} \label{sec:upweight}

\begin{figure}[t]
\centering
\hspace{-0.6cm}
\includegraphics[width=\linewidth]{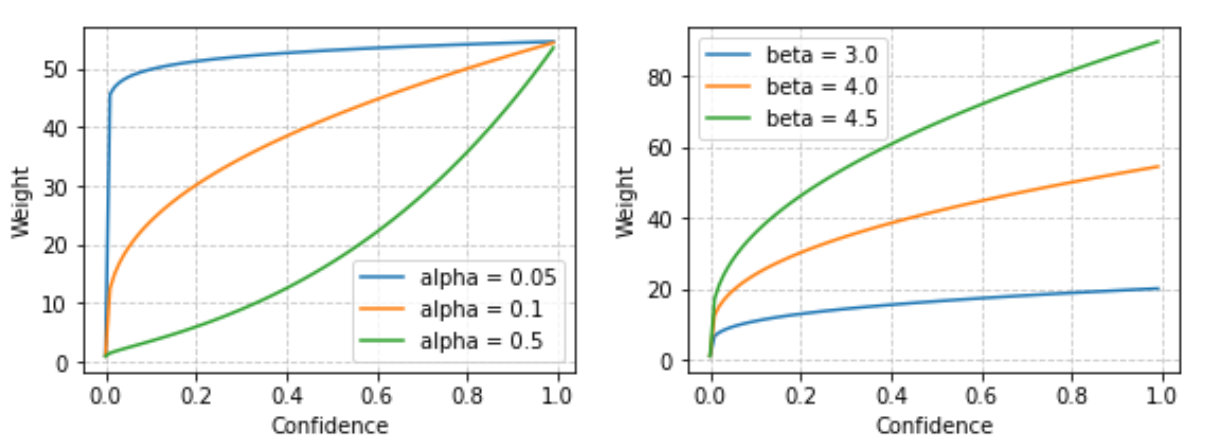}
\caption{Weighing scheme as a function of confidence of instances mispredicted by early readouts, for different values of $\alpha$ and $\beta$. a) shows weighing scheme for different values of $\alpha$ keeping $\beta$=4. b) shows weighing scheme for different values of $\beta$ keeping $\alpha$=0.1}
\label{fig:upweighting}
\end{figure}

In our experiments we observed that using mispredictions as the signal of an instance being from a worst group results in low-precision ($\sim$10-20\%) worst group identification. We counteract this by upweighing based on the confidence margin given \cref{fig:early_vs_late} (b, d)'s observation of higher confidence margin for worst group instances. This helps further increase the relative contribution of worst group instances to the loss, without using group labels.

\cref{fig:method-diagram} sketches the broad outline of our proposal. Let the representation learned by the neural network at a given depth $d$ be denoted by $g_d(x)$. In order to obtain early readouts, we add an auxiliary classifier layer $h_a$ on top of $g_{d}$. The early readouts are then the classifier probabilities $\mathbf{p}^{(aux)}= h_{aux}(g_d(x))$, with an associated predicted label $y^{(aux)}$, for a data sample $(x,y) \in D$. We train the new auxiliary network for $R$ epochs to obtain correct readouts. Then based on the early readouts, we devise a weighting scheme as follows:
\vspace{-0.1cm}
\begin{align}
\bm{wt}=\begin{cases}
			1, & \text{if } {y}^{({aux})} = y\\
             e^{\beta \cdot \mathtt{cm}(\mathbf{p}^{(aux)})^{\alpha}}, & \text{otherwise}
		 \end{cases}  
\label{eq:upweight}
\end{align}
where $\mathtt{cm}(\cdot)$ is the confidence margin described in \cref{eq:confmargin}. In other words, if the auxiliary prediction is correct, we do not change the instance weight; however, on erroneous predictions from the auxiliary network, we scale the associated (incorrect) confidence margin through an appropriately parametrized function to generate a weight.

The hyperparameters $(\alpha,\beta)$ control the variation of weights with confidence margin, and the maximum weight associated with any misclassified point, respectively. In \cref{fig:upweighting} (left) we present the influence of $\alpha$ when $\beta = 4$ is kept fixed; for lower $\alpha$ there is a sharp, substantial change in weights after a certain confidence value, and for higher $\alpha$ the changes are gradual. On the other hand, $\beta$ simply scales up the weight as a function of confidence for a given alpha (right panel). 

\subsection{Dynamic Reweighing for Distillation}
We augment the dataset $D$ with the weights $wt$ and obtain $\Dcal_w = \{(x_i,y_,wt_i)| i \in \{i \cdots n\}\}$. We perform knowledge distillation using a loss similar to the one described in Eq. \ref{eqdis1}, with our new augmented dataset $D_w$:
{\small
\begin{align}
      \mathcal{L}_{student} = \sum_{D_w}[ (1-\lambda) \cdot l_{ce} + \lambda \cdot \bm{wt} \cdot l_{kd}]
    \label{eq:final}
\end{align}}
where $wt$ are instance-specific weights as described above. We present the complete algorithm in \cref{alg:main}. Unlike previous approaches, we update weights $w_i$ throughout the training period (specifically, by retraining the auxiliary readout network $h_{aux}(\cdot)$ every $L$ epochs). This means that our weighting scheme is not only based on the properties of the model being trained (as opposed to a pretrained proxy model~\cite{liu2021just,lee2023debiased}), but also that it adapts during the training process. In particular, relative accuracies of different groups may change during training, and our approach seamlessly adapts the upweighting according to the needs of the current model ({\em c.f.}, \cref{sec:dynamicres}).

\algnewcommand\algorithmicforeach{\textbf{for each}}
\algdef{S}[FOR]{ForEach}[1]{\algorithmicforeach\ #1\ \algorithmicdo}
\newlength\myindent
\setlength\myindent{2em}
\newcommand\bindent{%
  \begingroup
  \setlength{\itemindent}{\myindent}
  \addtolength{\algorithmicindent}{\myindent}
}
\newcommand\eindent{\endgroup}
\begin{algorithm}[t]
\caption{The \ours{} approach: learning student $\mathcal{S}$, given dataset $\mathcal{D} = (x_i, y_i) \mid i \in (1 \cdots N)$ and teacher $\mathcal{T}$.}\label{alg:main}
\textbf{Hyperparameters:} Distillation loss fraction $\lambda$, depth $d$ of early readout, parameters $\alpha$ and $\beta$ for calculating the weights. 
\begin{algorithmic}[1]
\Statex
\State $\mathcal{S} = h(g_d(\cdot))$: break student down into early layers $g_d$ of depth $d$ and remaining deeper layers $h$. 
\State Let $h_{aux}(\cdot)$ be the auxiliary network 
\State We augment dataset $D$ with weights as
\Statex $\mathcal{D}_w \gets (x_i, y_i, wt_i)  \mid i \in (1 \cdots n)$, where $(wt_1 \cdots wt_N) \gets (1 \cdots 1)_n$
\Statex
\ForEach {$e \in \{1 \cdots E\}$}
    \State Train student model $\Scal$ using the loss described in \quad Eq. \ref{eq:final}.
    \If {$e \% L == 0$}: 
       \State Train $h_{aux}(g_d(\cdot))$ for $R$ epochs on dataset $D$ \quad \quad using standard cross-entropy. 
       \ForEach{$(x, y, wt) \in \mathcal{D}_w$}
          \State $y^{(\text{aux})} = h_{aux}(g_d(x)) $
          \State Update $wt$ according to Eq. \ref{eq:upweight}.
        \EndFor
    \EndIf
\EndFor
\end{algorithmic}
\end{algorithm}

\section{Experiment Setup}

\subsection{Datasets}\label{sec:datasets}

\begin{figure*}[t]
\centering
\hspace{-0.6cm}
\includegraphics[trim={0cm 6cm 0cm 0},clip,width=\linewidth]{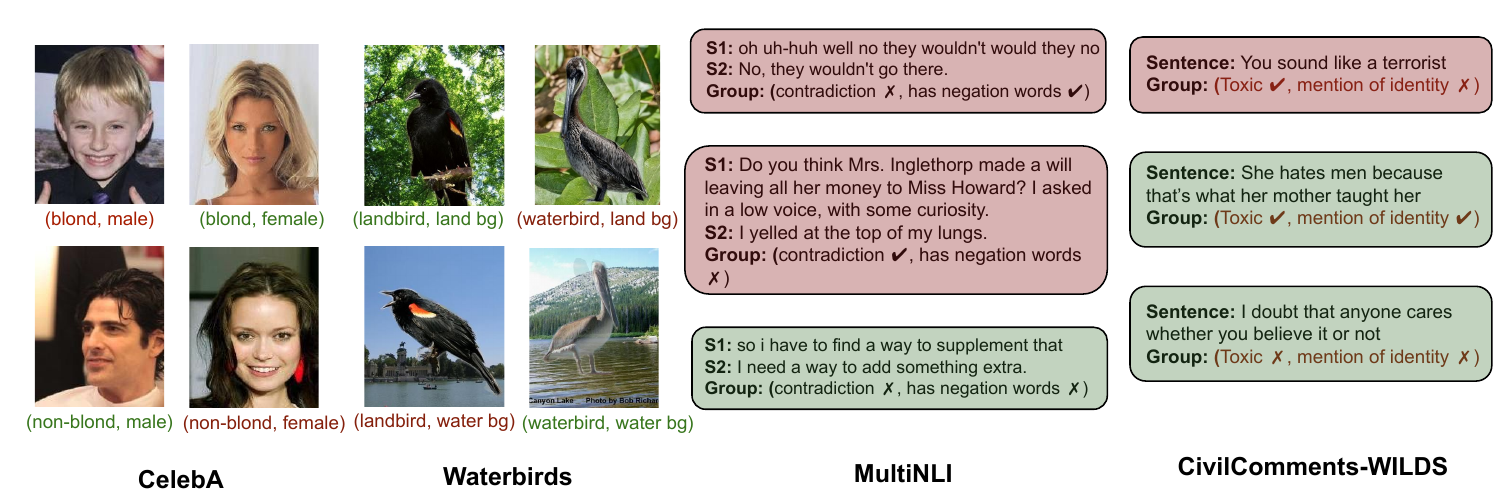}
\caption{Groups in the four datasets. Groups which follow the correlation are in green and ones in conflict with the correlation are in red.}
\label{fig:datasets}
\end{figure*}

We evaluate our method and various baselines on four debiasing benchmarks: \textit{Waterbirds}, \textit{CelebA}, \textit{MultiNLI} and \textit{CivilComments-WILDS}. Each of these are classification tasks where instances include a feature that is spuriously correlated with the label, as illustrated in Figure \ref{fig:datasets}. In other words, the feature or attribute value is often associated with specific labels without actually having any causal relationship with the label. This sets up a natural grouping of (label, attribute-value) pairs, where the attribute value may indicate a label that is consistent with, or in contrast to the true label. Empirically, models perform worst on groups with label-attribute conflict.

\noindent \textbf{Waterbirds~\cite{wah2011cub,sagawa2020group}:} Binary classification for bird images, with ``waterbird" and ``landbird" classes. Since each category is often pictured on a stereotypical background, the background (water, land) is spuriously correlated with the label. The dataset also contains small numbers of bias-conflicting examples for training and evaluation.

\noindent \textbf{CelebA\cite{liu2015deep}:} Binary classification of face images in order to identify hair color: ``blond" or ``non-blond"~\cite{sagawa2020group}. In the dataset, most blond-haired celebrities are female, setting up a spurious correlation between gender and hair color.

\noindent \textbf{MultiNLI\cite{williams2018broad}:} Each input sample consists of two sentences, and the task is to determine whether the meaning of the second sentence entailed by, neutral with, or contradicts the first sentence. The presence or absence of negation words in the second sentence can misleadingly influence the prediction task~\cite{sagawa2020group}.

\noindent \textbf{CivilComments-WILDS\cite{borkan2019nuanced,koh2021wilds}:} The objective is to categorize whether an online comment is toxic or not. In the dataset, the label is unintentionally correlated to references to specific demographic characteristics such as gender (male, female), ethnicity (White, Black), sexual orientation (LGBTQ), and religious affiliations (Muslim, Christian, and others). We use a binary indicator (appearance of words related to demographic identities in the comment) as the spuriously label-correlated attribute for defining groups.

\subsection{Evaluation Metric} 

We use the standard train, validation and test splits of each dataset. We assume that validation and test set have group information. As is common practice, we use group information of validation data for tuning hyperparameters, and group information of test data for measuring the performance of methods. We report \emph{Average Accuracy} and \emph{Worst Group Accuracy} (WGA) on the unseen test dataset. \emph{Average Accuracy} is the percentage of correctly predicted data points in the test set. For each dataset, test data is divided into groups as per \cref{sec:datasets}. Given these groups, WGA is the accuracy of the group with the worst performance for a given (method, dataset) combination.

\subsection{Baselines} \label{sec:baseline}

\begin{table*}[t]
\center
\resizebox{\textwidth}{!}{\renewcommand{\arraystretch}{1.2}
\begin{tabular}{lcccccccccc}
\toprule

\multicolumn{1}{c}{Method} &
  \multicolumn{1}{c}{Annotation} & \multicolumn{1}{c}{Teacher }&
  \multicolumn{2}{c}{Waterbirds} &
  \multicolumn{2}{c}{CelebA} &
  \multicolumn{2}{c}{CivilComments-WILDS} &
  \multicolumn{2}{c}{MultiNLI} \\ 
  \cmidrule(lr){4-5} \cmidrule(lr){6-7} \cmidrule(lr){8-9} \cmidrule(lr){10-11}
 &
  \multicolumn{1}{l}{} & &
  Avg Acc. &
  WGA &
  Avg Acc. &
  WGA &
  Avg Acc. &
  WGA &
  Avg Acc. &
  WGA \\
\hline  
  Teacher & No & No & 92.6 &	91.4 &	92.7 &	90.0 &	86.1 &	76.7 &	81.4 &	77.7 \\
\hline  
\rowcolor{lightgray}
  Group DRO & Yes & No &  87.0  {\footnotesize $\pm$ 1.65} &  81.1  {\footnotesize $\pm$ 1.60} &  93.0  {\footnotesize $\pm$ 0.31} &  86.4  {\footnotesize $\pm$ 1.27} &  82.2  {\footnotesize $\pm$ 0.80} &  77.3  {\footnotesize $\pm$ 0.81} &  50.7  {\footnotesize $\pm$ 1.73} &  47.5  {\footnotesize $\pm$ 1.33} \\
  \rowcolor{lightgray}
  ERM        & No  & No & 75.0  {\footnotesize $\pm$ 1.07} & 33.8  {\footnotesize $\pm$ 3.57} & \textbf{95.9}  {\footnotesize $\pm$ 0.12} & 44.1  {\footnotesize $\pm$ 1.41} & \textbf{91.0}  {\footnotesize $\pm$ 0.47} & 57.5  {\footnotesize $\pm$ 5.23} & 50.7  {\footnotesize $\pm$ 1.04} & 14.1  {\footnotesize $\pm$ 5.11}  \\
  \rowcolor{lightgray}
  JTT & No  & No & 86.7  {\footnotesize $\pm$ 0.50} & 80.5  {\footnotesize $\pm$ 0.53} & 86.4  {\footnotesize $\pm$ 4.65} & 77.8  {\footnotesize $\pm$ 2.48} & 84.0  {\footnotesize $\pm$ 3.21} & 60.0  {\footnotesize $\pm$ 2.06} & 55.1  {\footnotesize $\pm$ 0.89} & 25.2  {\footnotesize $\pm$ 3.44} \\
  \hline
  KD ($\lambda$ = 0.5) & No  & Yes & 87.9  {\footnotesize $\pm$ 0.90} & 67.9  {\footnotesize $\pm$ 0.68} & 95.6  {\footnotesize $\pm$ 0.25} & 62.0  {\footnotesize $\pm$ 4.51} & 87.5  {\footnotesize $\pm$ 2.94} & 69.1  {\footnotesize $\pm$ 6.47} & 57.5  {\footnotesize $\pm$ 0.50} & 46.2  {\footnotesize $\pm$ 0.17} \\

  KD ($\lambda$ = 1) & No  & Yes & 88.6  {\footnotesize $\pm$ 0.31} & 84.9  {\footnotesize $\pm$ 0.47} & 93.7  {\footnotesize $\pm$ 0.06} & 84.4  {\footnotesize $\pm$ 1.15} & 85.0  {\footnotesize $\pm$ 0.50} & 74.0  {\footnotesize $\pm$ 1.04} & 57.3  {\footnotesize $\pm$ 0.31} & 51.0  {\footnotesize $\pm$ 1.39} \\
  
  SimKD & No  & Yes & 82.1  {\footnotesize $\pm$ 1.10} & 71.4  {\footnotesize $\pm$ 2.15} & 93.0  {\footnotesize $\pm$ 0.31} & 89.0  {\footnotesize $\pm$ 0.35} & 87.0  {\footnotesize $\pm$ 0.40} & 74.0  {\footnotesize $\pm$ 2.25} & 70.7  {\footnotesize $\pm$ 0.92} & 64.1  {\footnotesize $\pm$ 1.62} \\
  
  DeTT & No  & Yes & 90.1  {\footnotesize $\pm$ 0.62} & 87.5  {\footnotesize $\pm$ 1.25} & 92.8  {\footnotesize $\pm$ 0.35} & 89.5  {\footnotesize $\pm$ 0.71} & 86.7  {\footnotesize $\pm$ 0.56} & 75.0  {\footnotesize $\pm$ 2.56} & 78.9  {\footnotesize $\pm$ 0.49} & 77.9  {\footnotesize $\pm$ 0.06} \\
\hline  
\ours\ (Ours) & No & Yes & \textbf{92.1}  {\footnotesize $\pm$ 0.39} &  \textbf{89.8}  {\footnotesize $\pm$ 0.47} &  93.2  {\footnotesize $\pm$ 0.06} &  \textbf{89.6}  {\footnotesize $\pm$ 1.67} &  84.4  {\footnotesize $\pm$ 0.39} &  \textbf{78.3} {\footnotesize $\pm$ 0.80} &  \textbf{80.1}  {\footnotesize $\pm$ 0.24} &  \textbf{78.9}  {\footnotesize $\pm$ 0.28} \\

\bottomrule
\end{tabular}
}
\vspace{-0.3cm}
\caption{Comprehensive comparison of methods across datasets. Rows represent various baselines, alongside the Teacher model and \ours\ (gray=supervised learning, white=distillation). Columns show average accuracy and worst group accuracy (WGA) on unseen test data, grouped by dataset. \ours{} substantially improves WGA compared to other distillation baselines, while simultaneously beating them on overall accuracy on 3 out of 4 datasets. We also consistently outperform Group DRO which uses group information in optimizing worst-group accuracy for a supervised setting.
}
\label{tab:main}
\end{table*}

Apart from standard knowledge distillation described in \cref{sec:kd}, we compare against the following:

    \noindent \textbf{Just Train Twice (JTT)} \cite{liu2021just}: This is designed for mitigating bias in supervised learning, without access to group information. The model is first trained for a small number of epochs \textit{n}, and misclassified instances are selected for follow-up interventions. A second classifier is trained from scratch with the loss from previously identified instances scaled up by a single fixed scalar $\lambda$. This method is limited by the dependence on tuning \textit{n} and $\lambda$, and lack of fine-grained distinction between instances. 

    \noindent \textbf{Group DRO} \cite{sagawa2020group}: This is also for supervised learning, with the use of both class and group labels during training. It aims to minimize the worst-group accuracy under early stopping. The need for annotations for group information is a major limitation; however, this method can be considered a skyline for methods that don't use group information.
    
    \noindent \textbf{KD with a reused teacher classifier (SimKD)} \cite{chen2022knowledge}: This is designed to improve knowledge distillation in general. This approach trains the student to mimic the final feature map (pre-classifier layer representation) from the teacher. Subsequently, the teacher's classification layer is grafted onto the student model, resulting in a much closer transfer of the teacher's knowledge. To address differences in output size between the teacher and student models, a projector layer is introduced following the student model's feature layer which leads to a small increase in the student's size. Though it does not specifically address debiasing, since larger teacher models are less biased, the better distillation mitigates bias to some extent. The limitations of this method are the need for changing the student architecture and increasing the student size.
    
    \noindent \textbf{Debiasing via Teacher Transplantation (DeTT)} \cite{lee2023debiased}: This is the baseline closest to our setting. It aims to mitigate bias, while simultaneously improving teacher matching, combining the approaches of both JTT and SimKD. They 1) use an unbiased teacher, 2) distil the teacher's feature map using a projection layer, and append the teacher classifier layer (from SimKD), 3) assign higher weights during training to initially misclassified samples (from JTT). The limitations of JTT and SimKD are inherited in DeTT.

\subsection{Model Architecture and Training} 
\label{sec: training_details}
We use the Resnet-18~\cite{he2016deep} architecture for the vision datasets, and  DistilBERT~\cite{sanh2019distilbert}  for the text datasets. The teacher model for knowledge distillation is Resnet-50~\cite{he2016deep} and BERT~\cite{kenton2019bert} for the vision \& text datasets respectively. We adopt a similar training approach as \cite{lee2023debiased}, training MultiNLI and CivilComments-WILDS datasets using AdamW \cite{loshchilov2018decoupled} and vision datasets using SGD optimizers. We do not use learning rate schedules. The learning rates, weight decay and position of the auxiliary layer, obtained after grid search, are listed in the appendix, along with the choice of hyperparameters $\alpha,\beta$. To reduce the search space, we keep the training interval ($L$) and number of training epochs for auxiliary layer ($R$) fixed at 1. The appendix also includes sensitivity analysis for the new hyperparameters. We train Waterbirds for 100, CelebA for 60, MultiNLI for 10 and CivilComments-WILDS for 10 epochs respectively. We employ validation worst group accuracy for early stopping.
We use a ResNet depth-1 block as the auxiliary network for vision datasets, and a two layer neural network for the text datasets. We train the auxiliary network at the end of each epoch. We implement \ours{} with $\lambda =1$.


\section{Results}

\cref{tab:main} compares \ouralgo{} on the standard benchmark datasets and the baselines described in Section \ref{sec:baseline}. We report \emph{Average Accuracy} and \emph{Worst Group Accuracy} (WGA). We aim to make significant gains in WGA, while ideally not worsening overall accuracy. 

\subsection{\ours: Debiased and Accurate}

\ours{} achieves substantial gains over baselines in \emph{Worst Group Accuracy} across datasets. An impressive result is that \ours{} also improves \emph{Average Accuracy} across different datasets (except for the CivilComments dataset), showing that our learned classifiers are overall more robust in addition to being less biased. 

This is because, unlike DeTT or JTT, we do not simply reweight a predetermined set of misclassified points that have been specified by some previous model; instead, \ours{} dynamically adapts the loss function through the mechanism of refreshing the early readout model every epoch. Thus, our classifiers are incentivized throughout the training process to focus on different problem instances, rather than an \textit{a priori} notion of worst group as per \cref{sec:datasets}. See \cref{sec:dynamicres} for further analysis of this key finding. We also observe that the earliest readouts give the best WGA improvements, refer to Appendix B.3 for more details. 

Note that \ours{} achieves better WGA than GroupDRO~\cite{sagawa2020group} which assumes availability of group information. For CivilComments and CelebA, we note that the debiasing baselines (JTT and DeTT) perform worse than ERM on average accuracy.


\subsection{Removing Spurious Correlations}
\begin{table}[t]
\centering
\resizebox{\columnwidth}{!}{\renewcommand{\arraystretch}{1.4}
{\begin{tabular}{c  cccc}
\toprule
Waterbirds groups & Teacher & DeTT & SimKD & \ours \\
\hline
  \rowcolor{lightgray}
(waterbird, water bg) & 94.3 & 92.6 {\footnotesize $\pm$ 0.70} & 89.4 {\footnotesize $\pm$ 0.06} & 94.1 {\footnotesize $\pm$ 0.86} \\
\rowcolor{lightgray}
(landbird, land bg) & 91.6 & 90.0 {\footnotesize $\pm$ 0.06} & 92.1 {\footnotesize $\pm$ 0.46} & \fbox{89.8} {\footnotesize $\pm$ 0.46} \\
(waterbird, land bg)  & 91.7 & \fbox{88.3} {\footnotesize $\pm$ 0.81} & \fbox{71.4} {\footnotesize $\pm$ 2.15} & 92.1 {\footnotesize $\pm$ 0.40} \\
(landbird, water bg) & \fbox{91.4 } & 88.8 {\footnotesize $\pm$ 1.70} & 84.6 {\footnotesize $\pm$ 0.79} & 90.6 {\footnotesize $\pm$ 0.67} \\

\hline
\vspace{3pt}
CelebA groups & &  &  &  \\
\hline
\rowcolor{lightgray}
(blond, female) & 94.3 & 92.6 {\footnotesize $\pm$ 1.14} & 92.2 {\footnotesize $\pm$ 0.46} & 92.7 {\footnotesize $\pm$ 1.48}  \\
\rowcolor{lightgray}
(non-blond, male) & 92.9 & 92.3 {\footnotesize $\pm$ 0.58} & 93.0 {\footnotesize $\pm$ 0.46} & 93.2 {\footnotesize $\pm$ 0.42}  \\
(non-blond, female) & 92.1 & 93.0 {\footnotesize $\pm$ 0.78} & 93.2 {\footnotesize $\pm$ 0.35} & 93.1 {\footnotesize $\pm$ 0.52}  \\
(blond, male) & \fbox{90.0} & \fbox{89.5} {\footnotesize $\pm$ 0.71} & \fbox{89.0} {\footnotesize $\pm$ 0.35} & \fbox{89.6} {\footnotesize $\pm$ 1.96} \\
\bottomrule 
\end{tabular}
}}

\center\small
\vspace{-0.3cm}
    \caption{Groupwise breakdown of accuracy for Waterbirds and CelebA datasets. Shown are the teacher model, \ours{} and KD baselines DeTT and SimKD. Gray rows indicate spurious feature values supporting label; white rows indicate conflict. Boxed text highlights worst-group performance in each column.}
\label{tab:groupwise-accs}
\vspace{-0.15cm}
\end{table}
We analyze the performance of \ours{}, alongside the Teacher model and baselines DeTT \& SimKD, across different subgroups in the held-out test set (\cref{tab:groupwise-accs}). The table distinguishes \textit{bias-confirming} groups where the spurious feature value \textit{supports} predicting the label (gray rows), and \textit{bias-contradicting} groups where the spurious feature value and label are in apparent conflict (white rows). For instance, in Waterbirds, a water background may (incorrectly) influence the classifier to predict a waterbird as label, and in CelebA, female gender may influence the classifier to predict blond hair color. These influences are helpful in the gray rows, and harmful in the white rows.

We note the following key findings: a) \ours{} beats baselines DeTT \& SimKD on not just worst group accuracy (\cref{tab:main}), but nearly all groups that have bias-conflicting feature values (\cref{tab:groupwise-accs}), and b) interestingly, our worst-group performance on Waterbirds is on a group that has a \textit{bias-confirming} combination of spurious feature and label. This confirms that our training procedure has successfully reduced the dependence on the spurious feature, without sacrificing overall accuracy, compared to baseline methods.

\subsection{Dynamics of Learning in \ours} \label{sec:dynamicres}
\begin{figure}[t]
\centering
\hspace{-0.6cm}
\includegraphics[width=0.95\linewidth]{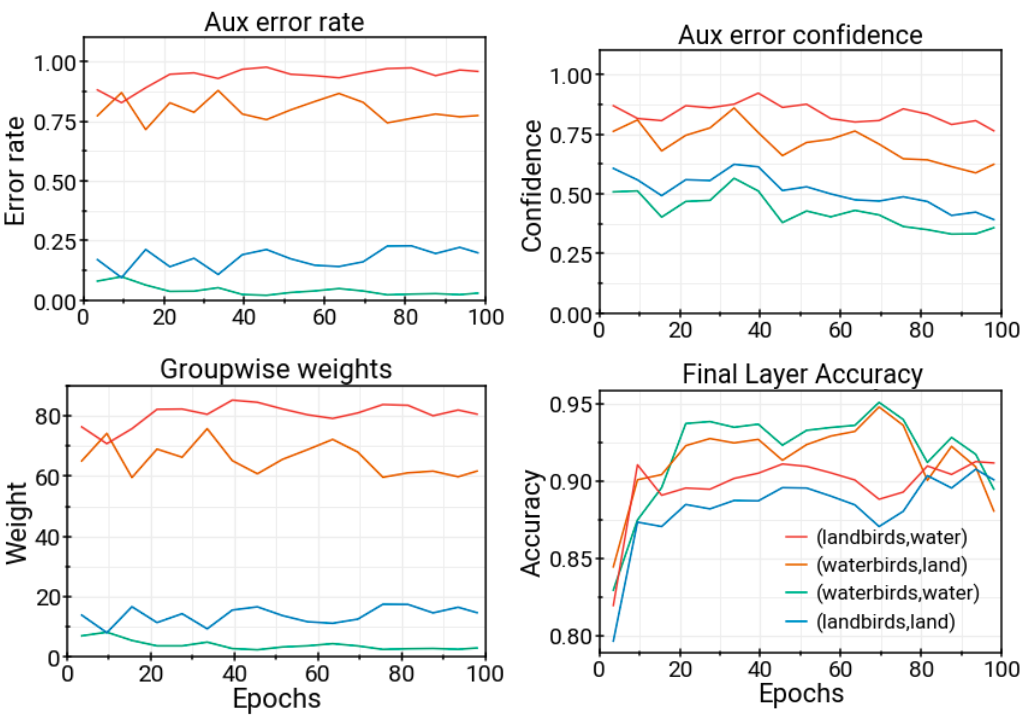}
\vspace{-0.1cm}
\caption{Evolution of reweighting during distillation (Waterbirds dataset). Top row shows error rate, and confidence of error instances, at the early readout broken down by groups. As expected, conflicting groups have high error rates due to spurious features; through the distillation process, the overconfidence reduces. Bottom row shows average weights for each group in the distillation loss (\cref{eq:final}), and the error rate at final layer. At the end of training, the groupwise accuracies are reconciled. }
\label{fig:analysis}
\vspace{-0.1cm}
\end{figure}

A key design aspect of \ours{} is the use of readouts \textit{from the model being trained} in order to refine the distillation loss (\cref{eq:final}), in contrast to JTT and DeTT. This means that the debiasing can dynamically vary throughout the training process to reflect the current state of the model. We illustrate these training dynamics in \cref{fig:analysis} for the Waterbirds dataset. We note the following interesting findings: 1) groups with spurious-feature/ label conflict have the highest early readout error (top left), and this is sustained through the training period, 2) the confidence associated with these errors reduces gradually over time (top right), 3) the average weights associated with each group changes as their performance (readout errors + confidence) changes at the auxiliary layer (bottom left v/s top left), 4) through the process of debiased distillation in \ours{}, conflicting and non-conflicting group accuracies are eventually reconciled, to be in close agreement (bottom right).

\section{Conclusion} 
We presented \ours{}, which debiases distillation by automatically identifying and mitigating student models' dependence on spurious correlations. We observe that labels decoded from earlier layers disproportionately and confidently fail on instances with conflict between label and spurious features. This novel finding shows that early readouts flag the risk of overdependence on spurious features, and leads naturally to a scheme for graded, per-instance adjustments to the distillation loss. We show via extensive experiments that \ours{} not only improves worst group accuracy but also overall accuracy across benchmark datasets, and that the adjustments to the distillation loss adapt to the evolution of the student model. 
In future work, we hope to explore the quality and robustness of features learnt by the student network, and the applicability of these ideas to broader settings such as domain generalization.




{\small
\bibliographystyle{ieee_fullname}
\bibliography{WACV}
}

\newpage
\appendix
\onecolumn
\leftline{ {\Large Appendix } }

\section{Additional Details}
\subsection{Group Composition in Datasets}
In \cref{sec:datasets} we introduce the various datasets used in our experiments. We present the group-wise distribution of data samples in Table \ref{tab:composition-breakdown}. 

\begin{table*}[h]
\centering
\small
\begin{adjustbox}{width=\linewidth}
{\begin{tabular}{cccccccc}
\toprule
Dataset & \multicolumn{7}{c}{Groupwise composition}\\
\toprule
\small
\multirow{4}{*}{Waterbirds} & &{\footnotesize (waterbird, water bg)} & {\footnotesize(waterbird, land bg)} &  {\fbox{\footnotesize(landbird, water bg)}}& {\footnotesize(landbird, land bg)} & & \\
&{\footnotesize Train} & {\footnotesize 72.7\%} & {\footnotesize 3.9\%} & {\footnotesize 1.2\%} &{\footnotesize 22.2\%} & &\\
&{\footnotesize Val}& {\footnotesize 38.9\%}& {\footnotesize 38.9\%} & {\footnotesize 11.1\%}& {\footnotesize 11.1\%} & & \\
&{\footnotesize Test}& {\footnotesize 38.9\%}& {\footnotesize 38.9\%} & {\footnotesize 11.1\%}& {\footnotesize 11.1\%} & & \\
\midrule
\multirow{4}{*}{CelebA} & &{\footnotesize (blond, female)} & {\footnotesize(non-blond, male)} &  {\footnotesize(non-blond, female) }& {\fbox{\footnotesize(blond, male)}} & & \\
 &{\footnotesize Train}& {\footnotesize 44.0\%}& {\footnotesize 41.1\%} & {\footnotesize 14.0\%} & {\footnotesize 0.9\% } & &\\
&{\footnotesize Val} & {\footnotesize 43.0\%} & {\footnotesize 41.7\%} & {\footnotesize 14.5\%} & {\footnotesize 0.9\%} & &\\
&{\footnotesize Test} & {\footnotesize 48.9\%} & {\footnotesize 37.7\%} & {\footnotesize 12.4\%} & {\footnotesize 0.9\%} & &\\
\midrule
\multirow{4}{*}{MultiNLI} & &{\footnotesize (contradiction, no-negation)} & {\footnotesize(contradiction, negation)} &  {\footnotesize(entailment, no-negation) }& {\fbox{\footnotesize(entailment, negation)}} & {\footnotesize(neutral, no-negation)}&{\footnotesize(neutral, negation)} \\
&{\footnotesize Train} & {\footnotesize 27.9\%} & {\footnotesize 5.4\%} & {\footnotesize 32.7\%} & {\footnotesize 0.7\%} & {\footnotesize 32.2\%} & {\footnotesize 1.0\%} \\
&{\footnotesize Val} & {\footnotesize 27.7\%}& {\footnotesize 5.6\%} &  {\footnotesize 32.7\%} & {\footnotesize 0.7\%} &{\footnotesize 32.3\%} & {\footnotesize 1.0\%} \\
&{\footnotesize Test} & {\footnotesize 28.0\%}& {\footnotesize 5.4\%} &  {\footnotesize 32.7\%} & {\footnotesize 0.7\%} &{\footnotesize 32.3\%} & {\footnotesize 0.9\%} \\
\midrule
\multirow{4}{*}{CivilComments} & &{\footnotesize (non-toxic, no-identity)} & {\footnotesize(non-toxic, identity)} &  {\fbox{\footnotesize(toxic, no-identity) }}& {\footnotesize(toxic, identity)} & & \\
&{\footnotesize Train} & {\footnotesize 53.4\%} & {\footnotesize 35.3\%} & {\footnotesize 4.4\%} & {\footnotesize 6.9\%} & & \\
&{\footnotesize Val} & {\footnotesize 53.9\%} & {\footnotesize 34.9\%} & {\footnotesize 4.4\%} & {\footnotesize 6.8\%} & &  \\
&{\footnotesize Test} & {\footnotesize 54.1\%} & {\footnotesize 34.5\%} & {\footnotesize 4.5\%} & {\footnotesize 6.8\%} & &  \\
\bottomrule 
\end{tabular}
}
\end{adjustbox}
\small
\vspace{-0.1cm}
    \caption{Different groups and their compositions in the training, validation and test splits of the four datasets. Boxed text highlights the minority group by frequency in training split--these groups by definition are ``in violation'' of the incidental / spurious feature correlation that is established by the majority group.}
\label{tab:composition-breakdown}
\end{table*}

\subsection{Training Details}
In this section we follow up on the details provided in the Section \ref{sec: training_details}. For all vision experiments, we consistently used ResNet-18 as the student model and ResNet-50 GroupDRO trained model as the teacher. For auxiliary layer we used 1 depth BasicBlock of ResNet. The ResNet network is composed of stages (each itself contains multiple BasicBlocks), we apply auxiliary layer only at the end of stages (except stage 4), this gives us three different choice for the hyperparameter aux position ($\mathcal{A}_P$). For text datasets, we used DistilBert as the student model and Bert GroupDRO trained model as the teacher. We used 2 layer neural network as an auxiliary layer that is applied at the end of encoder layers present in DistilBert.
Table \ref{tab:hparam-table} shows the hyperparameter search space for all the hyperparameters that we tune on the basis of worst group accuracy of validation set. Table \ref{tab:best-hyp-param} shows the best hyperparameter configurations choosen for each dataset from the result of this grid search.

\begin{table}[h]

\begin{center}
\begin{tabular}{ccc}
\toprule
\multicolumn{1}{c}{Hyperparameter} & Range &  \\ \midrule
$\alpha$                           & [0.01, 0.05, 0.1, 0.2, 0.5]    &  \\
$\beta$                            & [3, 3.5, 4., 4.5]          &  \\
$\mathcal{A_P}$                    & [1, 2, 3]                          & \\ 
$lr$                               & [1e-5, 2e-5, 5e-4, 1e-3]        & \\ 
$wd$                    & [0.1, 0.01, 0.001]                          & \\ \bottomrule

\end{tabular}
\end{center}
\caption{Range for hyperparameter search. Here $\alpha$ and $\beta$ are the weighting parameters, $\mathcal{A_P}$ determines the position of auxiliary layer, $lr$ is the learning and $wd$ denotes weight decay.}
\label{tab:hparam-table}
\end{table}

\begin{table}
\begin{center}
\begin{tabular}{lccccc}
\toprule
Dataset & $\alpha$ & $\beta$ & $\mathcal{A_P}$ &$lr$ & $wd$ \\
\midrule
Waterbirds       & 0.05           & 4      & 1       & 5e-4    & 0.1     \\
CelebA           & 0.1            & 3.5    & 1       & 5e-4    & 0.01        \\
CivilComments    & 0.05           & 3      & 1       & 2e-5    & 0.01      \\
MultiNLI         & 0.2            & 3      & 2       & 2e-5    & 0.01    \\
\bottomrule
\end{tabular}
\end{center}
\caption{Best hyperparameters for each dataset}
\label{tab:best-hyp-param}
\end{table}

\section{Additional experiments}
\subsection{Sensitivity Analysis}
In this section we show the sensitivity analysis of \ours{} on the Waterbirds dataset. Table \ref{tab:sensitivity} shows the average accuracy and worst group accuracy while distilling using \ours{} from GroupDRO teacher in a setting similar to Table \ref{tab:main} for different values of the weighting parameters $\alpha$ and $\beta$. We observe that tuning weighting parameters according to the dataset does help, but the performance is not too sensitive to these hyperparameters (as the standard deviation of the performance metric remains low across different hyperparameter values).
\begin{table}[h]
  \centering
    {\small{
\begin{tabular}{cc|cc}
\toprule
$\alpha$         & $\beta$       & Avg. Acc.     & Worst-group Acc. \\
\midrule
\textbf{0.05} & \textbf{4} & \textbf{92.3} & \textbf{90.3}    \\
0.05          & 3          & 91.5          & 89.7             \\
0.05          & 3.5        & 92.1          & 90.2             \\
0.05          & 4.5        & 91.1          & 90.3             \\
0.01          & 4          & 92.2          & 90.3             \\
0.1           & 4          & 91.1          & 88.9             \\
0.2           & 4          & 91.8          & 90.8             \\
0.5           & 4          & 91.1          & 89.5             \\ \midrule
              & Mean       & 91.6          & 90.0             \\
              & Std dev.   & 0.52          & 0.60             \\
\bottomrule
\end{tabular}
}}
\caption{Sensitivity analysis of \ours's performance to the weighting hyperparameters $\alpha$ and $\beta$ on the Waterbirds dataset.}
\label{tab:sensitivity}
\end{table}

\subsection{Additional Results}
\begin{figure}[h]
\centering
\hspace{-0.6cm}
\includegraphics[width=\linewidth]{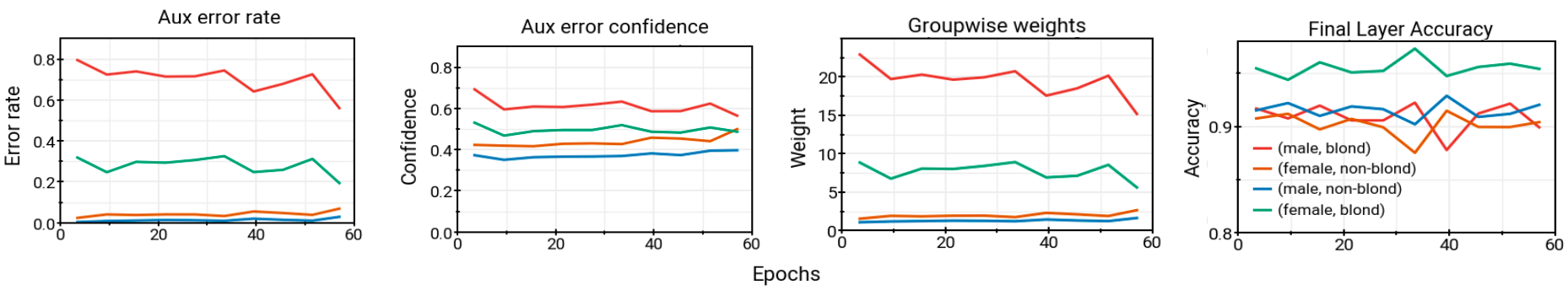}
\caption{Evolution of reweighting during distillation (CelebA
dataset). As expected, minority groups have high error rate; and through the distillation process, the overconfidence reduces}
\label{fig:analysis_celeba}
\end{figure}
We replicate Figure \ref{fig:analysis} in main paper for the training dynamics on the CelebA dataset with similar findings, thus showing that the dynamics of our method play a crucial role in training of the final classifier.

\subsection{Analyzing the importance of early readouts in \ours{}}
\begin{figure}[h]
\centering
\hspace{-0.6cm}
\includegraphics[width=0.5\linewidth]{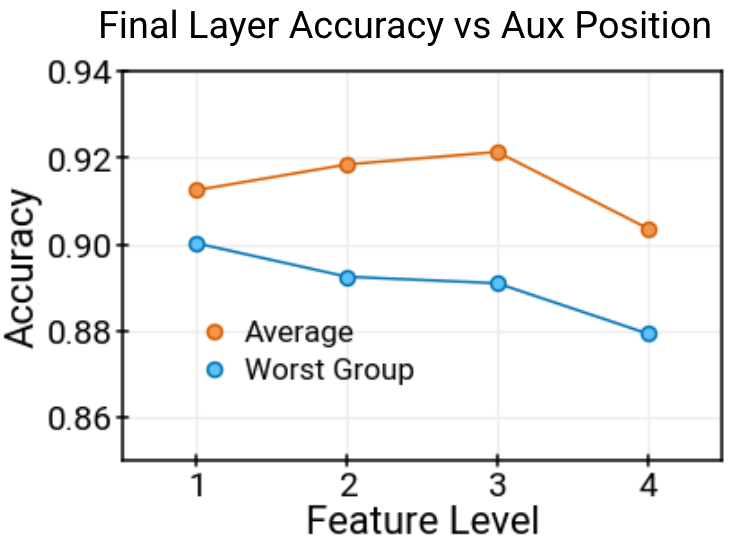}
\caption{Final performance on the Waterbirds dataset by \ours{} wrt feature layer chosen for readout. As expected, early readout leads to better debiased performance than using final layer.}
\label{fig:layer_analysis}
\end{figure}
To demonstrate the effect of \textit{early} readouts on the final performance, we show the results on Waterbirds dataset of performance on varying the position of auxiliary layer. As shown in figure \ref{fig:layer_analysis} the worst group accuracy shows a downward trend wrt to depth from which readout is taken. Overall, early-readouts (ie. feature level 1-3) are better than using the final layer (feature level 4), thus reaffirming the hypothesis mentioned in the main paper (Section 4.1). We have the same result on other datasets as well: ref. Table \ref{tab:best-hyp-param} where the optimum $\mathcal{A_P}$ is on the earliest layers.

\subsection{Analyzing weight assignments by \ours{}} 

For the datasets WaterBird and CelebA we analysed the weights assigned (at the end of training) and report the findings in Table \ref{tab:waterbirds-breakdown}.  In this table, we provide the mean weight for each group along with its standard deviation. Notably, we observe a discernible correlation between the assigned weights and the number of data points within each group. It is noteworthy that our method, \ours{}, tends to allocate larger weights to groups with fewer data points, effectively addressing the dataset's inherent imbalance. Remarkably, \ours{} accomplishes this without using label information to explicitly balance the dataset.

Furthermore, the weights detailed in Table \ref{tab:waterbirds-breakdown} underscore the effectiveness of our novel method for identifying the worst-performing group, even in the absence of explicit group information. It's apparent that the highest weights are consistently assigned to the group where the spurious correlations are broken, in both datasets. This finding further validates the improvements highlighted in Table \ref{tab:main}.

\begin{table}[h]
\centering
\resizebox{0.5\columnwidth}{!}{\renewcommand{\arraystretch}{1.4}
{\begin{tabular}{c  cc}
\toprule
Waterbirds' groups & $\#$Samples & Mean weight \\
\hline
(waterbird, water background) & 3498 & 3.2 ($\pm$ 3)	\\ 
(waterbird, land background)  & 184 & 39.07 ($\pm$ 6.73) \\
(landbird, water background) & 56 & 49.03 ($\pm$ 5.17)\\ 
(landbird, land background) & 1057 & 10.6 ($\pm$ 4.17) \\
\hline				
\vspace{3pt}
CelebA's groups & & \\ 
\hline
(blond, male) & 138 &19.52 ($\pm$ 3.37)	\\
(blond, female) & 22880	& 7.75 ($\pm$ 2.17)	\\	
(non-blond, male) & 66874 & 1.28 ($\pm$ 0.22)	\\
(non-blond, female) & 71629	& 1.97 ($\pm$ 0.5)	\\ 
\bottomrule 
\end{tabular}
}}

\small
\vspace{-0.1cm}
    \caption{Table presents the group distribution in training and the mean weightage given by \ours{} across the training. Note that minority groups are upweighted the most.}
\label{tab:waterbirds-breakdown}
\end{table}

\end{document}